\documentclass[10pt,twocolumn,letterpaper]{article}

\usepackage{cvpr}
\usepackage{times}
\usepackage{epsfig}
\usepackage{graphicx}
\usepackage{amsmath}
\usepackage{amssymb}
\usepackage{siunitx}
\usepackage[pagebackref=true,breaklinks=true,letterpaper=true,colorlinks,bookmarks=false]{hyperref}

\cvprfinalcopy % *** Uncomment this line for the final submission

\def\cvprPaperID{1461} % *** Enter the CVPR Paper ID here

\ifcvprfinal\fi
\usepackage{multirow}
\usepackage{booktabs}
\usepackage{pifont}% http://ctan.org/pkg/pifont
\usepackage{enumitem}

\newcommand{\cmark}{\ding{51}}%
\newcommand{\xmark}{\ding{55}}%

\newcommand{\zpd}[1]{{\color{black}{#1}}}
\newcommand{\zy}[1]{{\color{black}{#1}}}

\usepackage{amssymb}% http://ctan.org/pkg/amssymb
\usepackage[draft]{fixme}

\usepackage{soul}

\newcommand{\mn}[1]{{\color{black}{#1}}}
\newcommand{\mnh}[1]{{\color{black}{#1}}}

\usepackage{amsmath}
\usepackage{lipsum}

\ifcvprfinal\fi
\begin{document}
\title{Networks for Joint Affine and Non-parametric Image Registration}
\author{None}

\author{Zhengyang Shen \\UNC Chapel Hill\\{\tt\small zyshen@cs.unc.edu}
\and Xu Han\\UNC Chapel Hill\\{\tt\small xhs400@cs.unc.edu} \and Zhenlin Xu\\UNC Chapel Hill\\{\tt\small zhenlinx@cs.unc.edu} \and Marc Niethammer\\UNC Chapel Hill\\{\tt\small mn@cs.unc.edu}
}

\maketitle

\setlength{\abovedisplayskip}{3pt}
\setlength{\belowdisplayskip}{3pt}

\begin{abstract}
We introduce an end-to-end deep-learning framework for 3D medical image registration. In contrast to existing approaches, our framework combines two registration methods: an affine registration and a vector momentum-parameterized stationary velocity field (vSVF) model. Specifically, it consists of three stages. In the first stage, a multi-step affine network \mnh{predicts affine transform parameters.} In the second stage, we use a Unet-like network to generate \mnh{a momentum}, from which a velocity field can be computed via smoothing. Finally, in the third stage, we employ a self-iterable map-based vSVF \mnh{component} to provide \mn{a} non-parametric refinement based on the current estimate of the transformation map. Once the model is trained, a registration is completed in one forward pass. To evaluate the performance, we conducted longitudinal and cross-subject experiments on 3D magnetic resonance images (MRI) of the knee of the  Osteoarthritis Initiative (OAI) dataset. Results show that our framework achieves comparable performance to state-of-the-art medical image registration approaches, but it is much faster, with a better control of transformation regularity \mnh{including the ability to produce approximately symmetric transformations, and} combining affine and non-parametric registration.
\end{abstract}

\section{Introduction}

Registration is a fundamental task in medical image analysis to establish spatial correspondences between different images. To allow\mnh{, for example,} localized spatial analyses of cartilage changes over time or across subject populations, images are first registered to a common anatomical space. 

Traditional image registration algorithms, such as \mnh{elastic~\cite{bajcsy1989multiresolution,shen2002hammer}, fluid
~\cite{beg2005computing,hart2009optimal,vercauteren2009diffeomorphic,chen2013large,wulff2015efficient} or B-spline models~\cite{rueckert1999nonrigid}}, are based on the iterative numerical solution of an optimization problem. The objective of the optimization is to minimize image mismatch and transformation irregularity. The sought-for solution is then a spatial transformation which aligns a source image well to a target image while assuring that the transformation is sufficiently regular. %One of the most important metrics for registration quality is the similarity. Over the years a variety of different similarity measures to assess image mismatch have been proposed. For image pairs with a similar intensity distribution, Mean Square Error (MSE) on intensity differences is widely used. For multi-modal registration, however, Normalized Cross Correlation (NCC) and Mutual Information (MI) usually perform better. Besides, smooth transformation maps are typically desirable. 
To this end,  a variety of different similarity measures to assess image mismatch have been proposed. For image pairs with a similar intensity distribution, Mean Square Error (MSE) on intensity differences is widely used. For multi-modal registration, however, Normalized Cross Correlation (NCC) and Mutual Information (MI) usually perform better. Besides, smooth transformation maps are typically desirable. \mnh{Methods encouraging or enforcing smoothness use, for example,} rigidity penalties~\cite{staring2007rigidity} or penalties that encourage volume preservation~\cite{tanner2000volume,rohlfing2003volume} to avoid folds in the transformation. Diffeomorphic transformations can also be achieved by optimizing over sufficiently smooth velocity fields from which the spatial transformation can be recovered via integration. Such methods include Large Displacement Diffeomorphic Metric Mapping (LDDMM) ~\cite{beg2005computing,hart2009optimal} and Diffeomorphic Demons~\cite{vercauteren2009diffeomorphic}. As optimizations are typically over very high-dimensional parameter spaces, \mnh{they} are computationally expensive.

% However, as a common shortcoming of non-knowledge based registration methods, the transformation map is greedily searched and can be sensitive to local noise. Besides, optimization methods often need large number of iterations before convergence, with high demands on both time and computational resources. <- MN: This is not really a true statement. LDDMM or SVF are not greed algorithms for example. They optimize a clearly defined energy.

Recently, taking advantage of deep learning, research has focused on replacing costly numerical optimization with a learned deep regression model. These methods are extremely fast as only the evaluation of the regression model is required at test time.  
% A key aspect for non-parametric registration is what quantity to predict. In its simplest form, for instance, a deep network can directly predict a displacement field~\cite{rohe2017svf}. To assure spatial regularity, approaches to predict velocity fields or initial momenta fields have been proposed~\cite{yang2017quicksilver}. 
% While earlier work has focused on training models based on previously obtained registration parameters via \zpd{costly} numerical optimization~\cite{cao2018deformable,yang2016fast}, recent work has shifted to end-to-end formulations\footnote{For these end-to-end approaches, the sought-for registration parameterization is either the final output of the network (for the prediction of displacement fields) or an intermediate output (for the prediction of velocity fields) from which the transformation map can be recovered. The rest of the formulation stays the same.}~\cite{de2017end,li2017non,balakrishnan2018unsupervised,dalca2018unsupervised}. 
% %These end-to-end approaches integrate image resampling into their networks and hence were initially inspired by the spatial-transformer work of Jaderberg et al.~\cite{jaderberg2015spatial}.
% Both end-to-end approaches and non end-to-end approaches 
They imitate the behavior of conventional, numerical optimization-based registration algorithms as they predict the same types of registration parameters: displacement fields, velocity fields or momentum fields. Depending on the predicted parameters, theoretical properties of the original registration model can be retained. For example, In Quicksilver~\cite{yang2017quicksilver}, a network is learned to predict the initial momentum of LDDMM, which can then be used to find a diffeomorphic spatial transformation via LDDMM's shooting equations.  While earlier work has focused on training models based on previously obtained registration parameters via \zpd{costly} numerical optimization~\cite{cao2018deformable,yang2016fast}, recent work has shifted to end-to-end formulations\footnote{For these end-to-end approaches, the sought-for registration parameterization is either the final output of the network (for the prediction of displacement fields) or an intermediate output (for the prediction of velocity fields) from which the transformation map can be recovered. The rest of the formulation stays the same.}~\cite{de2017end,li2017non,balakrishnan2018unsupervised,dalca2018unsupervised}.  These end-to-end approaches integrate image resampling into their network \mnh{and were} inspired by the spatial-transformer work of Jaderberg et al.~\cite{jaderberg2015spatial}. Non end-to-end approaches require the sought-for registration parameters at training time. To obtain such data via numerical optimization for large numbers of image pairs can \mnh{be} computationally expensive, whereas end-to-end approaches effectively combine the training of the network with the implicit optimization over the registration parameters (as part of the network architecture).

%However, supervised learning requires ground truth parameters that obtained via conventional registration tools. Like the QuickSliver case, it sometimes takes long time to prepare the training data. More recently, many works ~\cite{de2017end,balakrishnan2018unsupervised,dalca2018unsupervised} focus on unsupervised strategies which refers to the idea of spatial transformer net~\cite{jaderberg2015spatial}.

%The common pipeline of those methods is designing networks for generating transformation map and then obtaining warped image via grid sampling. Though many works shows they perform comparable or better than conventional registration tools, no large boost has been reported. Compared with supervised learning method, unsupervised methods can be trained in an end2end way. Most unsupervised works rarely inherit the nice physical property like diffeomorphism from traditional methods.

Existing deep learning approaches to % non-parametric 
image registration exhibit multiple limitations. First, they assume that images have already been pre-aligned, \mnh{\eg}, by rigid or affine registration.  These \mnh{pre-alignment} steps can either be done via a specifically trained network \cite{chee2018airnet} or via standard numerical optimization. In the former case the overall registration approach is no longer end-to-end, while in the latter the \mnh{pre-registration becomes} the computational bottleneck. 
%\xuh{(I am a little skeptical about this sentence. Affine registrations on 3D images are not very slow and can be done within minutes. Reply: Yes, but networks often complete in seconds)} 
Second, many approaches are limited by computational memory and hence either only work in 2D or resort to small patches in 3D. Though some work explores end-to-end formulations for entire 3D volumes~\cite{balakrishnan2018unsupervised,dalca2018unsupervised}, these approaches perform computations based on the full resolution transformation map, in which case a very simple network can easily exhaust the memory and thus limit \mn{extensions} of the model. Third, they do not explore iterative refinement.

Our proposed approach addresses these shortcomings. Specifically, our contributions are:
\begin{itemize}[noitemsep]
  \item \mn{{\it A novel vector momentum-parameterized stationary velocity field registration model (vSVF)}}. The vector momentum field allows decoupling transformation smoothness and the prediction of the transformation parameters. Hence, sufficient smoothness of the resulting velocity field can be guaranteed and diffeomorphisms can be obtained even for large displacements.
  \item \mn{{\it An end-to-end registration method, merging affine and vSVF registration into a single framework}}. This framework achieves comparable performance to the corresponding optimization-based method and state-of-the-art registration approaches while dramatically reducing the computational cost.
  \item \mn{{\it A multi-step approach}} for the affine and the vSVF registration components in our model, which allows refining registration results.
  \item \mn{{\it An entire registration model via map compositions}} to avoid unnecessary image interpolations.
  \item \mn{{\it An inverse consistency loss both for the affine and the vSVF registration components}} thereby encouraging the regression model to learn a mapping which is less dependent on image ordering. \Ie, registering image A to B will result in similar spatial correspondences as registering \mn{B to A.}
  %\mnl{How novel is this in light of existing paper that have also proposed inverse consistency losses for registration using deep learning? Is there a difference in our formulation? Reply: Actually I didn't read any paper using affine and displacement constrains like this. But you may read some. Maybe to optimization methods, such kind of computation is too expensive.}
\end{itemize}

\begin{figure}[!t]
%\hspace*{-1cm}
\includegraphics[width=0.5\textwidth]{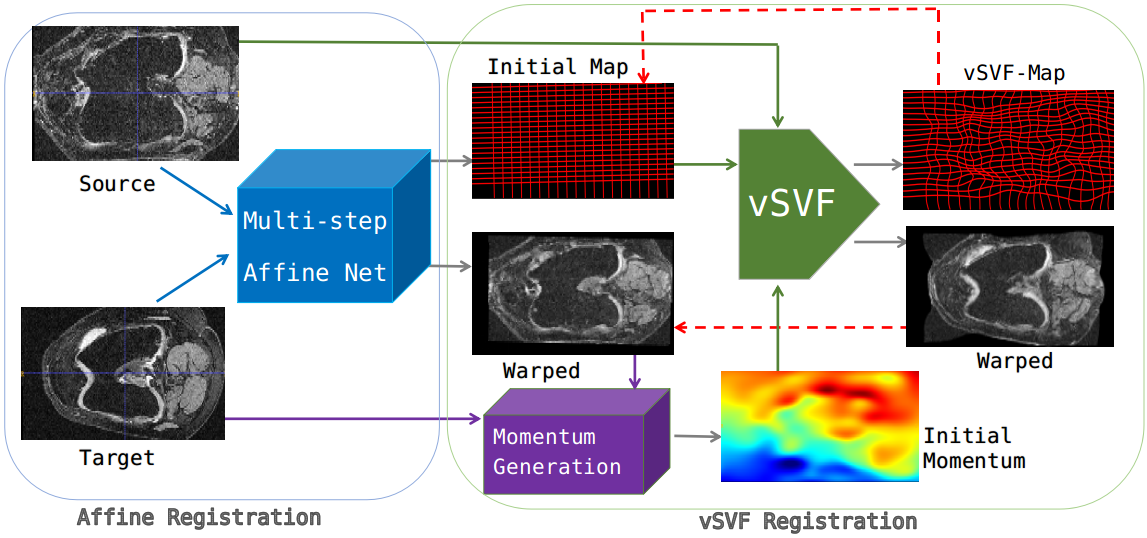}
\caption{Our framework consists of affine (left) and vSVF (right) registration components. The affine part outputs the affine map and the affinely warped source image. The \mnh{affine} map \mnh{initializes the} map of the vSVF registration. The affinely warped image \mnh{and} the target image \mnh{are} input into the momentum generation network to predict \mnh{the momentum} of the vSVF registration model. The outputs of the vSVF \mnh{component} are the composed transformation map and the warped source image, which can be either taken as the final registration result or fed back (indicated by the dashed line) into the vSVF \mnh{component} to refine the registration solution.}
%\mnl{I do not understand what the dashed lines are indicating. Please define.} \mnl{You sometimes use source and target, sometimes moving and target as terminology. Pick one and use it consistently}}
%\xuh{(I think the figure needs some refinement, very confusing for me to understand the pipeline. Why is the final warped image fed back to the middle warped image? Change font, add some arrows. Why are those blocks having different shapes? Also I think you should design all these figure as early as possible, as how much space is left for your writing is greatly dependent on these figures.)}
\label{fig:pipeline}
\end{figure}

Our approach facilitates image registration including affine pre-registration within one unified regression model. In what follows, we refer to our approach as AVSM (Affine-vSVF-Mapping). Fig.~\ref{fig:pipeline} shows an overview of the AVSM framework illustrating the combination of the affine and the vSVF registration components. 
%\xuh{(I suggest you put the figure in your method section.) Reply: Wait to discuss } 
The affine \mnh{and the vSVF components} are designed independently\mn{, but easy to combine.} %\mnh{Both components} support self-refinement.
\mn{In the affine stage}, a multi-step affine network predicts affine parameters for an image pair. In the vSVF stage, a Unet-like network generates \mnh{a momentum}, from which a velocity field can be computed via smoothing. The initial map and \mnh{the momentum} are then fed into the vSVF \mnh{component} \zpd{ to output the sought-for transformation map. A specified number of iterations can also be used to refine the \mnh{results}.} The entire registration framework operates on maps and uses map compositions. In this way, the source image is only interpolated once thereby avoiding image blurring. Furthermore, as the transformation map is assumed to be smooth, \mnh{interpolations to up-sample the map are accurate.} \mnh{Therefore, we can obtain good registration results by predicting a down-sampled transformation. However, the similarity measure is evaluated at full resolution during training}. Computing at low resolution greatly reduces the computational cost and allows us to compute on larger image volumes given a particular memory budget. \mnh{\Eg, a} map with \mnh{1/2 the size only requires 1/8 of the computations and 1/8 of the memory in 3D.} %\mnl{Is there an number we could put on how much more efficient it is with respect to computations? Reply: Fixed}
%\fixme{What do you mean by ``flexible with respect to image size''. In what way? Reply: Fixed. I rewrite this part}.

We compare AVSM to \mnh{publicly available optimization-based methods~\cite{ourselin2001reconstructing,modat2014global,rueckert1999nonrigid,modat2010fast,avants2009advanced}} \mnh{on longitudinal and \zpd{cross-subject} registrations} of 3D image pairs of the OAI dataset.

\mnh{The manuscript is organized as follows:} Sec.~\ref{sec:methods} describes our ASVM approach\zpd{;} Sec.~\ref{sec:experiments} \zpd{shows} experimental results\mnh{; Sec.~\ref{sec:conclusions_and_future_work}} presents conclusions and avenues for future work.

%To work on problems mentioned above, we propose an end-to-end map-based registration framework, which takes the advantage of the supervised learning but implement it in an unsupervised way. In specific, our model preserves all the physical property like diffeomorphism of the vector momentum-parameterized stationary vector field(vSVF), a novel fluid registration method we improve from stationary velocity field (SVF)~\cite{modat2012parametric}, and, at the same time, can be trained in an end2end way. Moreover, we build a multi-step affine network before the vSVF part. We design the model from map-based prospective, and will show its flexibility for multi-step registration and input image size. As the affine registration is assembled together with fluid based vSVF, transformation map can be predicted in a single pass, the framework realizes one-stop mapping. For easy notation, we name it as Affine-vSVF-Mapping (AVSM).

\section{Methods}
\label{sec:methods}

This section explains our overall approach. It is divided into two parts. The first part explains the affine registration component which makes use of a multi-step network to refine predictions of the affine transformation parameters. %apply linear and translation transform. %\fixme{You called this linear. I took it out. Only functions of the form y(x)=ax are linear. Something like y(x) = ax+c is affine. Reply: Sure, fixed}  
The second part explains the vector momentum-parameterized stationary velocity field (vSVF) which accounts for local deformations. Here, a momentum generation network first predicts \mnh{the momentum} parameterizing the vSVF model and therefore the transformation map. The vSVF \mnh{component} can also be applied in a multi-step way
%\fixme{Is self-iterable and multi-step something different for you? Reply: the same}
thereby further improving registration results.

\subsection{Multi-step Affine Network}

Most existing non-parametric registration approaches
%\footnote{\mnh{An exception} is curvature registration~\cite{modersitzki2004numerical} which uses second order differential operators for the regularizer, making it \mnh{affine invariant}.}
are not invariant to affine \mnh{transformations} as \mnh{they are penalized by the regularizers.} Hence, non-parametric registration approaches typically start from pre-registered image pairs, most typically based on affine registration, to account for \mnh{large, global} displacements or rotations. % during image acquisition. 
%Networks to predict affine registration parameters have been proposed~\cite{stergios2018linear}, but they are typically not integrated with networks for non-parametric registration.
%\cite{XXX}\fixme{I think I may have seen one such work, but I don't quite remember. Have you seen one?   Relpy:  Some work  on Arxiv did, but not published yet.  The paper titles are as follows: Linear and Deformable Image Registration with 3D CNNs   A Deep Learning Framework for Unsupervised Affine and Deformable Image Registration}.
%Hence, overall registration networks are either not end-to-end or (if the affine registration is done via numerical optimization) fast for the non-parameteric registration part but comparatively slow for the initial pre-registration. Our goal is to create a network that integrates affine transformation and the non-parametric registration model (based on vSVF). Here, we describe the affine component which % (inspired by the spatial transformer work~\cite{jaderberg2015spatial}
%\fixme{How is it inspy the spatial transformer work? Reply: I remove this part, but actually the multi-step affine first comes out of that paper, though not the same composed way like ours}) %
Therefore, in the first part of our framework, \mnh{we} use a multi-step affine network directly predicting the affine registration parameters and the corresponding transformation map.

The network needs to be flexible enough to adapt to both small and large affine deformations. Although deep convolutional networks can have large receptive fields, our experiments show that training a single affine network does not perform well in practice.
%\xuh{(Is there any such experiment?) Rely: Yes, in ablation study}
\mnh{Instead, we compose the affine transformation from several steps.} This strategy \mn{results in} significant improvements in accuracy and stability.

\begin{figure}[!t]
\centering
\includegraphics[width=0.3\textwidth]{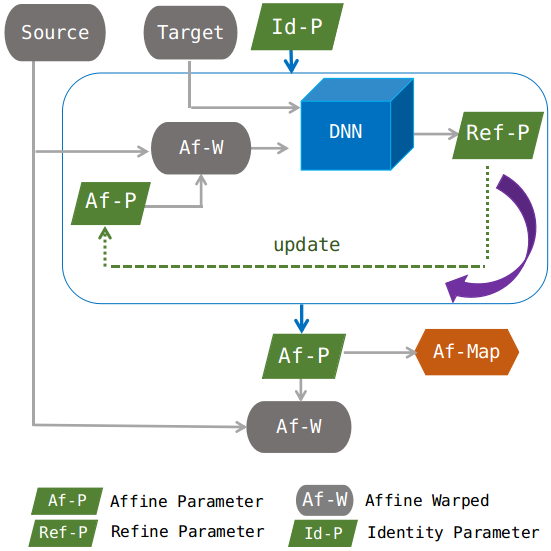}
\caption{\label{fig:affine} \mnh{Multi-step affine network structure}. As in a recurrent network, the parameters of the affine network are shared by all steps.
At \mnh{each step}, the network outputs the parameters to refine the previously predicted affine transformation. \mnh{\Ie,} the current estimate is obtained by composition (indicated by dashed line). The overall affine transformation is obtained at the last step.%\mnl{I would suggest renaming these blocks to: Affine transformation, identity transformation, refinement transformation, ... Reply: It may be a little bit confusing to using transformation here, I modify the image a little bit to see if it better explains.}
}
\end{figure}

\vskip1ex
\noindent
\textbf{Network}: Our multi-step affine network is a recurrent network, which progressively refines the predicted affine transformation.
%\xuh{(What do you mean by  sharing the same parameters? Are they different networks or just one network that you run it multiple times? What's the different between yours vs RNN? Reply: Yes,same as rnn.Multi-step is more reasonable for registration task)}
Fig.~\ref{fig:affine} shows the network architecture.
To avoid numerical instabilities and numerical dissipation due to successive trilinear interpolations, we directly update the affine registration parameters rather than resampling images in intermediate steps.
%\mnl{TODO!!!!!!      Do you have experimental evidence that shows that such instabilities happen? Reply: The multi-step methods do not work well in image based case. We can put it into appendix}. 
Specifically, at each step we take the target image and the warped source image (obtained via interpolation from the source image using the previous affine parameters) as inputs and then output the new affine \mnh{parameters} for the transformation refinement. Let the affine parameters be $\Gamma=\begin{pmatrix} A & b\end{pmatrix}$ , where $A \in \mathbb{R}^{d\times d}$ represents the linear transformation matrix; $b\in \mathbb{R}^d$ denotes the translation and $d$ is the image dimension. The update rule is as follows:
%\mnr{In scientific papers it is standard practice to number all equations. In this way reviewers can easily reference them. Reply: Sure}
\begin{equation}
\begin{split}
  &A_{(t)}= \tilde{A}_{(t)}A_{(t-1)},\;b_{(t)}= \tilde{A}_{(t)}b_{(t-1)} + \tilde{b}_{(t)},\\& s.t. \quad A_{(0)}=I,~b_{(0)}=0.
 \end{split}
\end{equation}
%\xuh{(Did you mention any where $I$ is the identity matrix?)Reply: Yes}
Here, $\tilde{A}_{(t)}$, $A_{(t)}$ represent the linear transformation matrix output and the composition result at the \mnh{$t\text{-}$th step}, respectively.
%\xuh{(You can use $A_t$ and just say the composition result?)} 
Similarly, $\tilde{b}_{(t)}$ denotes the affine translation parameter output at the $t\text{-}$th step and $b_{(t)}$ the composition result. Finally, if we consider the registration from the source image to the target image in the space of the target image, the affine map is obtained by \mnh{$\Phi_a^{-1}(x,\Gamma ) = A_{(t_{last})}x+b_{(t_{last})}$.}

\vskip1ex
\noindent
\textbf{Loss}: The loss of the multi-step affine network consists of three parts: an image similarity loss $L_{a \text{-}sim}$, a regularization loss $L_{a \text{-}reg}$ and a loss encouraging transformation symmetry $L_{a \text{-}sym}$. Let us denote $I_0$ as the source image and $I_1$ as the target image. The \mnh{superscripts} $^{st}$ and $^{ts}$ denote \mnh{registrations} from $I_0$ to $I_1$ and $I_1$ to $I_0$, respectively\footnote{To simplify the notation, we \mnh{omit $^{st}$ (source to target registration) in what follows and} only emphasize $^{ts}$ (target to source registration).}. 

%\mnl{It would be good to specify in the equation what these terms depend on. $I_0\circ\Phi^{-1}$, $I_1$???, Reply: I try to add but it seems sym loss includes too many terms would make the equation very long. so I make a footnote, let $^{st}$  denote the registration direction from target to source}

\mnh{The \textbf{image similarity loss $L_{a\text{-}sim}(I_0,I_1,\Phi_a^{-1})$}} can be any standard similarity measure, \eg, Normalized Cross Correlation (NCC), Localized NCC (LNCC), or Mean Square Error (MSE). Here we generalize LNCC to a multi-kernel LNCC formulation (mk-LNCC). Standard LNCC is computed by averaging NCC scores of overlapping sliding windows centered at \mnh{\emph{sampled} voxels}. Let $V$ be the volume of the image; $x_i$, $y_i$
%\mnl{Why do you switch to x and y here if you sued I0 and I1 before? This is confusing} 
refer to the $i^{th}$ ($i \in \{1,..,\mnh{|V|}\}$) voxel in the warped source and target volumes, respectively. $N_s$ refers to the number of sliding windows with cubic size $s\times s \times s$. %\mnl{What does size s mean? Is it a cube?} 
%\xuh{(I added warped source image and target image) Rely: Sure}
Let $\zeta_j^s$ refer to the window centered at the $j^{th}$ voxel and $\bar{x}_j,\bar{y}_j$ to the average image intensity \mn{values} over $\zeta_j^s$ in the warped source and target image, respectively. \mnh{LNCC} with window size $s$, denoted as $\kappa_s$, is defined by
%\mnr{You need to be much more careful and precise with your mathematical equations. Many of them are not really proper. I tried to fix them as I could. But please check that what I rewrote them as is in fact what you meant and what you implemented. Rely: Sure, I would be careful}
\begin{equation}
  \kappa_s(x,y) = \frac{1}{N_s}\sum_j{\frac{\sum\limits_{i \in \zeta_j^s}{(x_i-\bar{x}_j})(y_i-\bar{y}_j)}{\sqrt{\sum\limits_{i \in \zeta_j^s}{(x_i-\bar{x}_j})^2\sum\limits_{i \in \zeta_j^s}{(y_i-\bar{y}_j})^2}}}\enspace.
\end{equation}
%\mnl{This is not accurately describing what you are implementing is it? I thought you use some form of striding and $i$ does not range across all voxels, does it? Fix this. Please be precise in all your equations otherwise this is simply incorrect. And please also specify arguments of equations if possible, e.g., $\kappa_s(x,y)$ instead of just $\kappa_s$.}
\mn{We define} mk-LNCC as a weighted sum of LNCCs with different window sizes. For computational efficiency LNCC can be evaluated over windows centered over a subset \mn{of} voxels of $V$. The image similarity loss \mnh{is then}
\begin{equation}
\begin{split}
&L_{a \text{-}sim}(I_0,I_1,\Gamma )= \sum_i{\omega_i\kappa_{s_i}(I_0\circ \Phi_a^{-1}, I_1)}, \\&\text{s.t.}\mkern15mu \Phi_a^{-1}(x,\Gamma ) = Ax+b\mkern10mu\text{and}\mkern10mu\sum_i{\omega_i} = 1, w_i\geq 0\mnh{.}
\end{split}
\end{equation}

\mnh{The \textbf{regularization loss $L_{a \text{-}reg}(\Gamma )$} penalizes deviations} of the composed affine transform from \mnh{the identity:} 
\begin{equation}
  L_{a \text{-}reg}(\Gamma ) = \lambda_{ar}(||A-I||_F^2 + ||b||_2^2),
\end{equation}
where $\|\cdot\|_F$ denotes the Frobenius norm and \mnh{$\lambda_{ar}\geq 0$} \mnh{is an epoch-dependent weight factor designed to be large at the beginning of the training to constrain large deformations and then gradually decaying to zero}. See Eq.~\ref{eq:regularization_penalty_weights} for details.
%\xuh{(Are the $A$ and the $b$ here the composite result from all previous steps or the current output of the network?) Reply: fixed, I change  notation  a little bit}

\mnh{The \textbf{symmetry loss $L_{a \text{-}sym}(\Gamma , \Gamma ^{ts})$}} encourages the registration to be inverse consistent. \mnh{\Ie,}, we want to \mnh{encourage} that the transformation computed from source to target image is the inverse of the transformation computed from the target to the source image \mnh{(\ie, $A^{ts}(Ax+b)+b^{ts} = x$):}
%\Longrightarrow A^{ts}A=I, A^{ts}b+b^{ts}=0
%where $ts$ indicates target to source and $st$ source to target. This then motivates the loss function
\begin{equation}
  L_{a \text{-}sym}(\Gamma , \Gamma ^{ts}) = \lambda_{as}(||A^{ts}A-I||_F^2 + ||A^{ts}b+b^{ts}||_2^2),
\end{equation}
% where $A$,~$A^{ts} $are the linear matrices 
% %\xuh{(Again, I think $A$ can not be called affine matrix), Reply: fixed}
% learned for source-target and target-source separately and $b$, $b^{ts}$ are the corresponding translation vectors. 
where $\lambda_{as}\geq 0$ is a chosen constant.
%\mnl{As for the previous penalty, do you indeed not use square penalties here? Also, I changed the norm to the Frobenius norm. Is this what you did?Reply: yes, you are right}

%Before training the vSVF part of the network, we train the affine network separately. 

\mnh{The \textbf{complete loss $\mathcal{L}_a(I_0,I_1,\Gamma ,\Gamma ^{ts})$ } is then:}
%\mnr{Mathematical equations are part of sentences. Almost always there needs to be a comma or period at the end. Rely: Got it}
\begin{equation}
  \begin{split}
  \mathcal{L}_a(I_0,I_1,\Gamma ,\Gamma ^{ts}) =&\ell_a(I_0,I_1,\Gamma )+ \ell_a(I_1,I_0,\Gamma ^{ts}) \\&+ L_{a \text{-}sym}(\Gamma , \Gamma ^{ts}),
  \end{split}
\end{equation}
where $\ell_a(I_0,I_1,\Gamma ) = L_{a \text{-}sim}(I_0, I_1,\Gamma) + L_{a \text{-}reg}(\Gamma)$.
% \mnl{This equation is still not clear. What is $\Gamma $ for example? And why not simply define the loss as: $l_a(I_0,I_1,\Phi_a^{-1}) := L_{a-sim}(I_0\circ\Phi_a^{-1} I_1)$, then you can write the symmetrized version as something like: $l_a(I_0,I_1,\Phi_a^{-1}) + l_a(I_1,I_0,(\Phi_a^{ts})^{-1})$ or something similar. These terems need to have arguments: $l_a(...)$?, $\mathcal{L}_a(...)$?}

\subsection{Vector Momentum-parameterized \mnh{SVF}}
This section \mnh{presents the momentum based stationary velocity field method followed by the network to predict the momentum. For simplicity, we describe the one step vSVF here, which forms the basis of the multi-step approach.}

\vskip1ex
\noindent
\textbf{vSVF Method}: To capture large deformations and to guarantee diffeomorphic transformations, registration algorithms \mnh{motivated} by fluid mechanics are frequently employed. Here, the transformation map $\Phi$ \footnote{The subscript ${_v}$ of $\Phi_{v}$ is omitted, where \textit{v} refers to vSVF method.} in source image space is obtained via time-integration of \mnh{a velocity field $v(x,t)$, which needs to be estimated}. \mnh{The governing differential equation is:} $\Phi_t(x,t) = v(\Phi(x,t),t),~\Phi(x,0)=\Phi_{(0)}(x)$, where $\Phi_{(0)}$ is the initial map. For \mnh{a sufficiently smooth velocity field $v$ one} obtains a diffeomorphic transformation. Sufficient smoothness is achieved by penalizing non-smoothness of \mnh{$v$. Specifically, the} optimization problem is
\begin{equation}
\begin{split}
  v^* = &~\underset{v}{\text{argmin}}~\lambda_{vr} \int_0^1 \|v\|_L^2~\mathrm{d}t + \text{Sim}[I_0\circ\Phi^{-1}(1),I_1], \\\text{s.t.}\quad
  & \Phi^{-1}_t + D\Phi^{-1}v=0\quad\text{and}\quad\Phi^{-1}(0)=\Phi^{-1}_{(0)}\enspace\mnh{,}
\end{split}
\raisetag{20pt}
\end{equation}
%\xuh{(What is $\lambda_r$ here? Is it the same one in the loss section? If not, a different notation is needed.) Reply: Fixed}
\mnh{where} $D$ denotes the \mnh{Jacobian and} $\|v\|^2_L=\langle L^\dagger L v,v\rangle$ is a spatial norm defined \mnh{by specifying} the differential operator $L$ and its adjoint $L^\dagger$. %Picking a specific $L$ implies picking an expected model of deformation. In its simplest form, such a differential operator is \emph{spatially-invariant} and encodes a desired level of smoothness.
As the vector-valued momentum $m$ is \mnh{equivalent to} $m=L^\dagger L v$, one can \mnh{express} the norm \mnh{as} $\|v\|_L^2 = \langle m,v\rangle$. In the LDDMM approach~\cite{beg2005computing}\mnh{, time-dependent vector fields $v(x,t)$ are estimated}. A slightly simpler approach is \mnh{to use a} \emph{stationary velocity field} (SVF) $v(x)$ ~\cite{modat2012parametric}.
%\xuh{(Are you citing the correct one? They proposed an svf-net not the svf itself.) Reply: Fixed}
The rest of the formulation remains the same. While the SVF registration algorithms optimize directly over the velocity field $v$, we propose a 
\emph{vector momentum SVF (vSVF)} formulation which is computed as
\begin{equation}
\begin{split}
    &m^* = ~\underset{m_0}{\text{argmin}}~\lambda_{vr}\langle m_0,v_0\rangle + 
    \text{Sim}[I_0\circ\Phi^{-1}(1),I_1], \mkern2mu\text{s.t.}\\ &\Phi^{-1}_t + D\Phi^{-1}v=0,\Phi^{-1}(0)=\Phi^{-1}_{(0)}, v_0=(L^\dagger L)^{-1}m_0,
  \label{eq:vsvf}
\end{split}
\raisetag{25pt}
\end{equation}
where $m_0$ denotes \mnh{the vector} momentum and \mnh{$\lambda_{vr}>0$} is a \mnh{constant.} This formulation can be \mn{considered a simplified} version of the vector momentum-parameterized LDDMM formulation~\cite{vialard2012diffeomorphic}. The benefit of such a formulation is that it allows us to explicitly control spatial smoothness as the deep network predicts the momentum which gets subsequently smoothed to obtain the velocity field, instead of predicting the velocity field $v$ directly which would then require the network \mnh{to learn to} predict a smooth vector field. % We introduce and use vSVF for its simplicity, but note that our approach directly translates to LDDMM and in fact is motivated by the desire to obtain LDDMM regularizers which can adapt to a deforming image.

\begin{figure}[!t]
\includegraphics[width=0.5\textwidth, height=4cm]{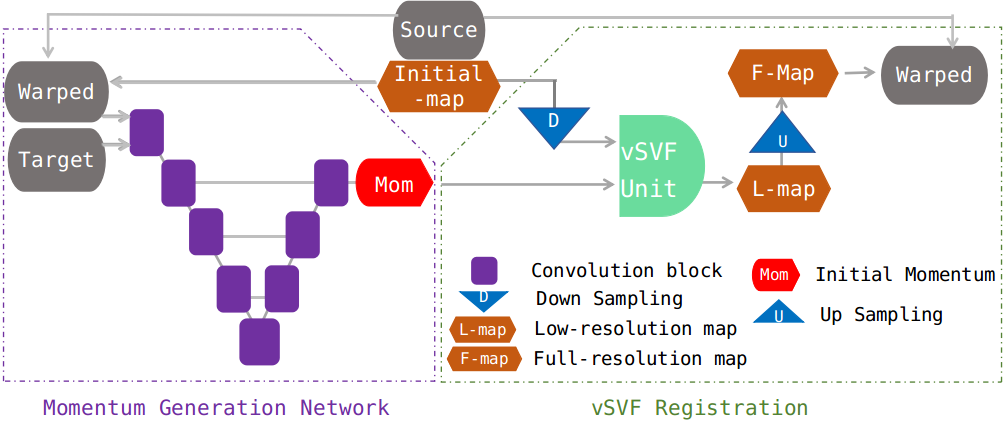}
\caption{\label{fig:vSVF} \mnh{vSVF registration framework illustration (one step)}, including the momentum generation network and the vSVF registration. The network outputs a low-resolution momentum. The momentum \mnh{and} the down-sampled initial map \mnh{are input to the vSVF \zy{unit} outputting a low-resolution transformation map}, which is then up-sampled to full resolution before warping the source image.
% The network outputs a low-resolution momentum , which is smoothed to obtain the corresponding velocity field. The vSVF unit  smoothed the momentum then integrates along time steps the velocity field to obtain a low-resolution transformation map, which is then up-sampled to full resolution before warping the source image.
}
\end{figure}

 Fig.~\ref{fig:vSVF} illustrates the framework of the vector momentum-parameterized stationary velocity field (vSVF) registration.  We compute using a low-resolution velocity \mnh{field, which} greatly reduces memory consumption. The framework consists of two parts: 1) a momentum generation network \mnh{taking} as the input \zy{the warped source image}, together with the target image, outputting the \mnh{low-resolution momentum}\mnh{;} 2) the vSVF registration part. 
 \mnh{Specifically,} the predicted momentum \mnh{and the} down-sampled initial map \mnh{are input into the vSVF \zy{unit}}, the output of which is finally up-sampled to obtain \mnh{the} full resolution transformation map.  Inside the vSVF \zy{unit}, a velocity field is \mnh{obtained by smoothing the} momentum and then \mnh{used to solve the} advection equation, $ {\Phi^{-1}_{(\tau)t}} + D\Phi^{-1}_{(\tau)}v=0$, for \mnh{unit time (using several discrete time points)}\mnh{. This then results in the sought-for transformation map.} The initial map mentioned here can be the affine map or the map obtained from \mnh{a previous vSVF step}, namely for the $\tau$-th step, set $\Phi_{(\tau)}^{-1}(x,0)=\Phi_{(\tau-1)}^{-1}(x,1)$. %

\vskip1ex
\noindent
\textbf{Momentum Generation Network}: We implement a deep neural network to generate the vector momentum. As our work does not focus on the network architecture, we simply implement a four-level U-net with residual \mnh{links~\cite{ronneberger2015u,milletari2016v}. Implementation details can be found in the supplementary material.} During training, the gradient is first backpropagated through the integrator for the advection equation followed by the momentum generation network. This can require a lot of memory.
%\mnl{I do not understand this? First of all is the gradient not always back-propagated? Second, what is the difference between the solution of the advection equation (this is not really shooting here) and the momentum network? I tried to fix it, but am not sure if I have done this correctly. Please check.} 
Therefore, to reduce memory requirements, the network outputs a low-resolution momentum. In practice, we remove the last decoder level of the U-net. In this case, the remaining vSVF \mnh{component} also operates on the low-resolution map. %which can significantly save the memory when registering image pairs of large size.

\vskip1ex
\noindent
\textbf{Loss}: Similar to the loss in the affine network, the loss for the vSVF part of the network also consists of three terms: a similarity loss  $L_{v \text{-}sim}$, a regularization loss $L_{v \text{-}reg}$ and a symmetry loss $L_{v \text{-}sym}$.

\mnh{The \textbf{similarity loss} $L_{v \text{-}sim}(I_0,I_1,\Phi^{-1})$} is \mnh{the same as for the affine network. \Ie, we also use mk-LNCC.}%: $\sum_i{\omega_i\kappa_{s_i}(I_0\circ \Phi^{-1}, I_1)}$.

\mnh{The \textbf{regularization loss} $L_{v \text{-}reg}(m_0)$} penalizes the velocity field. Thus, we have
\begin{equation}
    L_{v \text{-}reg}(m_0) = \lambda_{vr}\|v\|_L^2 =\lambda_{vr} \langle m_0,v_0\rangle, 
\end{equation}
where $v_0=(L^\dagger L)^{-1}m_0$. \mnh{We implement $(L^\dagger L)^{-1}$ as a convolution with a} multi-Gaussian kernel~\cite{risser2010simultaneous}.

\mnh{The \textbf{symmetric loss} is defined as}
\begin{equation}
    L_{v \text{-}sym}(\Phi^{-1},(\Phi^{ts})^{-1})= \lambda_{vs}\|\Phi^{-1}\circ(\Phi^{ts})^{-1}- id\|_2^2,
\end{equation}
where $id$ denotes the identity map, $\lambda_{vs}\geq 0$ refers to the symmetry weight factor, $(\Phi^{ts})^{-1}$ denotes the map obtained from registering the target to the source image in the space of the source image and $\Phi^{-1}$ denotes the map obtained from registering the source image to the target image in the space of the target image.  Consequentially, the composition also lives in the target image space. 

\mnh{The \textbf{complete loss $\mathcal{L}_v(I_0,I_1,\Phi^{-1},(\Phi^{ts})^{-1},m_0,m_0^{ts})$}} for vSVF registration with one \mn{step} is as follows:
\begin{equation}
\begin{split}
\mathcal{L}_v(I_0,I_1,&\Phi^{-1},(\Phi^{ts})^{-1},m_0,m_0^{ts}) = \ell_v(I_0,I_1,\Phi^{-1},m_0)\\&+ \ell_v(I_1,I_0,(\Phi^{ts})^{-1},m_0^{ts})\\&+ L_{v\text{-}sym}(\Phi^{-1},(\Phi^{ts})^{-1}),
%\text{s.t.}\quad&\ell_v(I_0,I_1,\Phi^{-1},m_0) = L_{v \text{-}sim}(I_0,I_1,\Phi^{-1})\\
%&+L_{v \text{-}reg}(m_0)
\end{split}
\raisetag{10pt}
\end{equation}
where:\\
$\ell_v(I_0,I_1,\Phi^{-1},m_0) = L_{v \text{-}sim}(I_0,I_1,\Phi^{-1})  +L_{v \text{-}reg}(m_0)$.

For the vSVF model with $T$ steps, the complete \mnh{loss is:}
\begin{equation}
\begin{split}
\sum_{\tau=1}^{T}\mathcal{L}_v(&I_0,I_1,\Phi_{(\tau)}^{-1},{\Phi_{(\tau)}^{ts}}^{-1},m_{0(\tau)},m^{ts}_{0(\tau)}) \quad \text{s.t.}\\
 & \Phi_{(\tau)}^{-1}(x,0)=\Phi_{(\tau-1)}^{-1}(x,1),\\
&(\Phi_{(\tau)}^{ts})^{-1}(x,0)=(\Phi_{(\tau-1)}^{ts})^{-1}(x,1)\mnh{.}
\end{split}
\raisetag{30pt}
\end{equation}
% \Phi_{(0)}^{-1}&=\Phi_{(0)}\\
% (\Phi_{(0)}^{ts})^{-1}&= \Phi_{(0)}^{ts}.
%\xuh{Add two equations with regards to $\Phi^{-1}_{(0)}$, $\Phi^{-1}_{(0)} = id?$ Reply: I have already mentioned that before. The initial map? $\Phi_{(0)}$ can be any initial map}
%Why do you open 4 overleaf sessions?

\section{Experiments and Results}
\label{sec:experiments}

%\mnh{Make sure all the experimental details are in here so that one can replicate the results. Some of the material you have in the methods description are in my opinion more appropriate for the experiments section. For example, the description of the penalty weights that change with the epochs, e.g., $\lambda_r$.}

\noindent
\textbf{Dataset}: The Osteoarthritis Initiative (OAI) dataset %\footnote{oai.epi-ucsf.org} 
consists of 176 manually labeled \mnh{magnetic resonance (MR)} images from 88 patients (2 longitudinal scans per patient) and 22,950 unlabeled MR images from 2,444 patients. Labels are available for femoral and tibial cartilage. All images are of size $384\times384\times160$, where each voxel is of size  $0.36\times0.36\times0.7mm^3$. We normalize the intensities of each image such that the $0.1$th percentile and the $99.9$th percentile are mapped to ${0,1}$ and clamp values that are smaller to $0$ and larger to $1$ to avoid outliers. All images are down-sampled to size $192\times 192\times 80$. 

%We further resample all \mnh{affinely} 
%\mnl{This is correct, right? You wrote linearly aligned, but I don't know what linear alignment is.} 
%aligned images to have the same spatial resolution (\mnh{\ie, }$1mm\times1mm\times1mm$).\mnl{??Size is 192x192x80??}
%\zy{actually, I found some paper descirbe like that,  but I think we just modifty the spacing while not resample anything}
%\xuh{(Is the spacing doubled, after you downsampled the image?) Yes, but we further do some alignment, align all spacing into [1,1,1]}
%\mnl{Is this a good strategy? But I guess it is not that important if you use NCC, I directly use the data that zhenlin processed before} 
%To compare with baseline methods limited by computational resources, .

\vskip0.5ex
\noindent
\textbf{Evaluation}: We evaluate on both longitudinal and cross-subject registrations. We divide the unlabeled patients into a \mnh{training and a} validation group, with a ratio of 7:3. %The patients with labels form the test group.
For the \textit{longitudinal registrations}, 4,200 pairs from the training group \mn{(obtained by swapping the source and the target from 2,100 pairs of images)} are randomly selected for training, and 50 pairs selected from the validation group are used for validation. All 176 longitudinal pairs with labels are used as our test set. For the \textit{cross-subject registrations}, we randomly pick 2,800 (from 1,400 pairs) cross-subject training pairs and 50 validation pairs; 300 pairs (from 150 pairs) are randomly selected as the test set. We use the average Dice score~\cite{dice1945measures} over all testing pairs as the evaluation metric.

\vskip0.5ex
\noindent
\textbf{Training details}: The training stage includes two parts:

%\begin{itemize}
%\item[1)] 
{\it 1) Training multi-step affine net:} It is \mnh{difficult} to train \mnh{the} multi-step affine network from scratch.
Instead, we train a single-step network first and use its parameters to initialize the multi-step network. \mnh{For longitudinal registration, we} train with a three-step affine network, but use a seven-step network during testing. This results in better testing performance than a three-step network. Similarly, \mnh{for cross-subject registration} we train with \mnh{a} five-step network and test with \mnh{a} seven-step one. \mnh{The affine} symmetry factor $\lambda_{as}$ is set to 10.

%\item[2)] 
\
{\it 2) Training momentum generation network:} During training, the affine part is fixed. For vSVF, we use 10 time-steps and a multi-Gaussian kernel with standard deviations \{0.05, 0.1, 0.15, 0.2, 0.25\} and corresponding weights \{0.067, 0.133, 0.2, 0.267, 0.333\} (spacing is scaled so that the image is in $[0,1]^3$). 
%\xuh{(I find it hard to understand, maybe spacing is scaled so the largest dimension of the image is 1?)Reply: actually in my implementation all dimensions are in [0,1]} 
We train with two steps for both longitudinal and cross-subject registrations.
%\mnl{This all seems a bit arbitrary, affine: 3/5, here 2/6, ...} 
\mnh{The} vSVF regularization factor $\lambda_{vr}$ is set to 10 and the symmetry factor $\lambda_{vs}$ is set to \zy{1e-4}. % I check the code, using mean rather than sum
%\end{itemize}
For both parts, we use the same training strategy: 1 pair per batch, 400 batches per epoch, 200 epochs per experiment;  we set a learning rate of 5e-4 with a decay factor of 0.5 after every 60 epochs. We use mk-LNCC as the similarity measure with \mnh{$(\omega,s)=\{ (0.3, S/4), (0.7, S/2)\}$}, where $S$ \mnh{refers} to \mnh{the} smallest image dimension. Besides, in our implementation of mk-LNCC, we set the sliding window stride to $S/4$ and kernel dilation to 2.

Additionally, the affine regularization factor $\lambda_{ar}$ \mnh{is {\it epoch-dependent} during training and defined as:}
\begin{equation}
  \lambda_{ar} \mnh{:=} \frac{C_{ar} K_{ar}}{K_{ar}+e^{n/K_{ar}}}\mnh{,}
\label{eq:regularization_penalty_weights}
\end{equation}
where $C_{ar}$ \mnh{is a constant}, $K_{ar}$ controls the decay rate\mnh{, and} $n$ is the epoch count. In both longitudinal and cross-subject experiments, $K_{ar}$ is set to 4 and  $C_{ar}$ is set to 10.
%\xuh{(Is $\lambda_{vs}$ no longer time-dependent?) Reply: yes, it is set as a constant}
% put into appendix
 %Specifically, at the beginning of the affine training stage, the output of the network is random and thus requires a strong regularization to force the output parameters to a reasonable range (close to identity). As a result, the weight $\lambda_{ar}$ is large at the beginning of the training to constrain the deformation. After several epochs, we observed that the training process stabilizes and no longer requires this penalty. Hence, we use the decay function. Similarly, for $\lambda_vs$, during the vSVF training, we do not strongly penalize asymmetry until the model parameters are already roughly optimized.

% \begin{figure}[!h]
%     \centering
%     \subfloat[]{{\includegraphics[width=5cm]{Figs/decay_factor.png} }}%
%     \qquad
%     \subfloat[]{{\includegraphics[width=5cm]{Figs/increase_factor.png} }}%
%     \caption{Illustration for epoch-dependent regularization penalty, $\lambda_r$. (a) The sigmoid decay function restrains the affine deformation in affine registration. (b) The sigmoid explode function increase the weight of symmetric loss in SVF registration.\mnl{I cannot follow what you want to say here. What is a ``sigmoid explode'' function for example? Please add short descriptions to all captions that summarize what I am supposed to see here. Also, please make all labels of figures readable, i.e., with reasonable font sizes.}}%
%     \label{fig:factor}%
% \end{figure}

\begin{figure}[!h]
%\hspace*{-0.7cm} 
\includegraphics[width=0.5\textwidth]{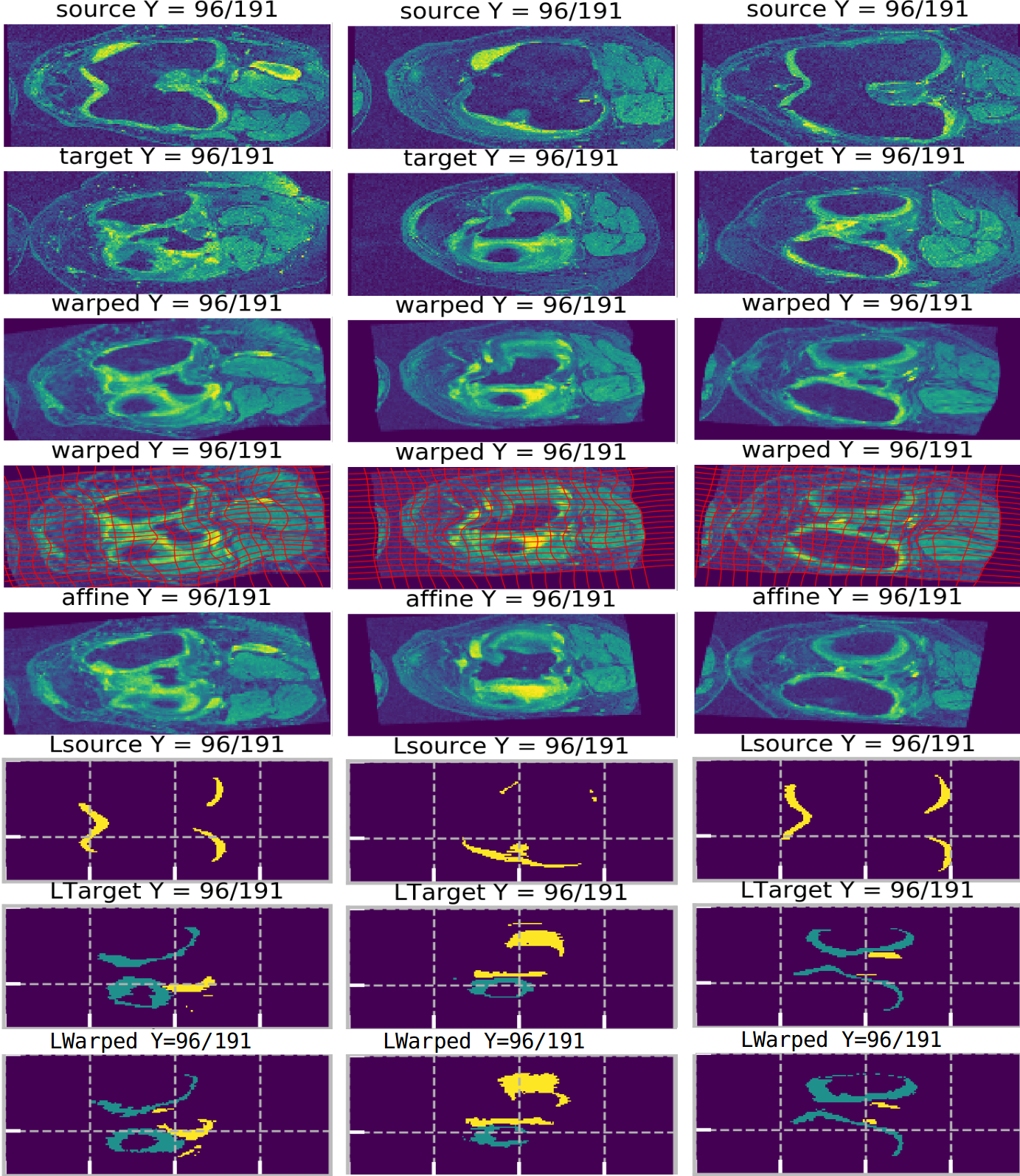}\\
\caption{\label{fig:example} Illustration of registration results achieved by AVSM, each column refers to an example. The first five rows refer to source, target, warped image by AVSM, warped image \mn{with deformation grid (visualizing $\Phi^{-1}$)}, warped image by multi-step affine respectively, followed by source label, target label and warped label by AVSM separately. \mnh{There is high similarity between the warped and the target labels and the deformations are smooth.}
}
\end{figure}

%\vskip0.5ex
\noindent
\textbf{Baseline methods}: We implement the corresponding \mnh{numerically} optimized \mnh{versions (\mnh{\eg,} directly optimizing \mnh{the momentum)}} of affine (affine-opt) and vSVF (vSVF-opt) registrations. We compare with three widely-used public registration methods: \textit{SyN} ~\cite{avants2009advanced,avants2008symmetric}, \textit{Demons} ~\cite{vercauteren2009diffeomorphic,vercauteren2008symmetric} and \textit{NiftyReg} ~\cite{ourselin2001reconstructing,modat2014global,rueckert1999nonrigid,modat2010fast}. Besides, we also compare the most recent VoxelMorph variant~\cite{dalca2018unsupervised}. \mnh{We report} their performance after \mnh{an in-depth} search \mnh{for good parameters}. \mn{For Demons, SyN and NiftyReg, we use isotropic voxel spacing $1\times 1\times 1mm^3$ as this gives improved results compared to using physical spacing. This implies anisotropic regularization in physical space. For our approaches, isotropic or anisotropic regularization in physical space gives similar results. Hence, we choose the more natural isotropic regularization in physical space.}
%The detailed settings of the public methods are listed in \textit{Appendix}.

%For all the baseline methods, \mnh{we report} their performance after \mnh{an in-depth} search \mnh{for good parameters}. We find \mnh{that} resetting the image spacing \mnh{to} $[1mm\times 1mm \times 1mm]$ can largely improves the performance of public tools\mnh{is this true for all of them: Syn, NiftyReg and demons? Also, I talked to Zhenlin and he told me he only changed the origin and not the spacing. Are you sure you are using uniform spacing?}, so we report their results based on this setting\footnote{\mnh{Note that the} physical spacing \mnh{of the} OAI \mnh{images} is \mnh{highly} asymmetric. \mnh{As the} smoothers are commonly defined \mnh{in} physical space \mnh{this} would cause asymmetric smoothing in image space.\mnl{And why is this a bad thing? Is it?} Though it is common to do registration in physical space, in our case the image-coordinate symmetric strategy works better.\mnl{It is really not clear why this would be the case. Do you actually want more regularity in the out-of-plane dimension.}}.
%\mnl{Needs to be clarified (potentially in supplementary material) how these methods were run; i.e., what settings, deformation model, similarity measure, ...}

\vskip0.5ex
\textit{Optimization-based multi-scale affine registration}: Instead of optimizing for the affine parameters on a single image scale, we use a multi-scale strategy. %In practice, directly optimizing affine parameters according to high resolution image space easily goes into wrong direction and collapse the model.
Specifically, we start at a low image-resolution, where affine parameters are roughly estimated, and then use them as the initialization for the next higher scale. Stochastic gradient descent is used with a learning rate of 1e-4. Three image scales $\{0.25, 0.5, 1.0\}$ are used, each with $\{200, 200, 50\}$ iterations. \mnh{We use mk-LNCC as the similarity measure.} At each scale \mnh{$k$}, let image size (smallest length among image dimensions)
%\mnl{You need to define what you mean by the $S_k$ at the very beginning when you define your mk-LNCC measure; you also need to motivate somewhere why you do not want to use LNCC.} 
be $S_k$, here $k \in \{0.25,0.5,1.0\}$. \mnh{At scale 1.0}, parameters are set to $(\omega, s)=\{  (0.3, S_k/4), (0.7, S_k/2)\}$, \mnh{\ie, the same parameters as for} the network version; at \mnh{scales} 0.5 and 0.25, $(\omega, s)=\{ (1.0, S_k/2)\}$.

\vskip0.5ex
\textit{Optimization-based multi-scale vSVF registration}: We take the affine map (resulting from the optimization-based multi-scale affine registration) as the initial map and then numerically optimize the vSVF model. The same multi-scale strategy as for the affine registration is used.
%At low-scale, svf captures large displacements and at the top-scale, it fine-tunes the local deformation.\mnl{Did you observe this? I don't think this is entirely true. It really depends on what the downsampled images look like. At all scales the model is the same, just the images it sees are a bit different. Reply: make sense, I remove this sentence} 
The \mnh{momentum is up-sampled between scales}. \mnh{We use L-BGFS~\cite{liu1989limited} for optimization.} In our experiments, we \mnh{use} three scales $\{0.25, 0.5, 1.0\}$ with 60 iterations per scale. The same mk-LNCC similarity measure as for the \textit{optimization-based multi-scale affine registration} is used. \mn{The number} of time steps for the integration of the advection equation and the settings for the multi-Gaussian kernel are the same as for the proposed deep network model.
%\mnl{It would be better to state at the beginning for which models all these settings are the same.}\mnl{Is this initialized with the affine results from above? You need affine pre-registration for vSVF.}

\textit{NiftyReg}: %Developed by Centre for Medical Image Computing at University College London,
We run two registration phases: affine followed by B-spline registration.
%\mnl{What does with embedded functions mean?}
Three scales are used in each phase and the interval of the B-spline control points is set to 10 voxels. 
%\mnl{I thought you observed better results with a smaller spacing?} 
In addition, we find that using LNCC as the similarity measure, with a standard deviation of 40 for the \mnh{Gaussian} kernel, performs better than the default Normalized Mutual Information\mnh{, but introduces folds in the transformation}. In LNCC experiments, we \mnh{therefore use a log of the Jacobi determinant penalty of $0.01$} to reduce folds.
%\mnl{What does tet mean here? There is no training phase is there?}, 
%20-thread is used for acceleration.
%\mnl{Why do you only give these computational details here? Describe them in one spot for all of these algorithms if possible.}
% why dd <- Must have been a typo, does not mean anything

\textit{Demons}: \zy{We take the affine map obtained from NiftyReg as the initial map and use the \textit{Fast Symmetric Forces Demons Algorithm} \cite{vercauteren2009diffeomorphic} via SimpleITK}. The Gaussian smoothing standard \mn{deviation} for the displacement field is \mnh{set to 1.2. We use MSE as the similarity measure.}

\textit{SyN}: We compare with Symmetric Normalization (SyN), a widely used registration method implemented in the ANTs software \mn{package~\cite{avants2008symmetric}}. We take Mattes as the metric for affine registration, and take CC with sampling radius set to 4 for SyN registration. \mnh{We use multi-resolution optimization with four scales with \{2100, 1200, 1200, 20\} iterations; the standard deviation for Gaussian smoothing} at each level is set to \{3, 2, 1, 0\}\mnh{. The flow standard deviation to smooth the gradient field is set to 3.} 

\textit{VoxelMorph}: We compute results based on the most recent VoxelMorph variant~\cite{dalca2018unsupervised}, which is also a deep-learning based. As VoxelMorph assumes that images are pre-aligned, for a fair comparison, we therefore initialized it via our proposed multi-step affine network. Best parameters are determined via grid search.

\textit{NiftyReg}, \textit{Demons} and \textit{SyN} are \mnh{run} on a server with i9-7900X (10 cores @ 3.30GHz)
%\mnl{Are these 20 threads? Or what server is this?}
, while all other methods run on a single NVIDIA GTX 1080Ti.

\begin{table}[!t]
\centering
%\hspace{-1cm}
\scalebox{0.68}{
\tabcolsep=0.11cm
\begin{tabular}{|cccccc|}
  %\toprule
  \hline
\multirow{2}{*}{Method} &
\multicolumn{2}{c}{Longitudinal} &
\multicolumn{2}{c}{Cross-subject} &
\multicolumn{1}{c|}{} \\
& Dice & Folds & Dice & Folds& Time (s) \\\hline%\midrule

affine-NiftyReg   &75.07 (6.21)  &0 & 30.43 (12.11) &0  & 45\\
affine-opt  &  \textbf{78.61} (4.48) &0 & 34.49 (18.07) &0  & 8\\     
affine-net (7-step)  &77.75 (4.77) &0 & \textbf{44.58} (7.74) &0  & 0.20  \\

--------------& & & & &\\
Demons     & \textbf{83.43} (2.64) &10.7 [0.56]  & 63.47 (9.52) &19.0 [0.56]&114\\
SyN        & 83.13 (2.67) &0  & 65.71 (15.01) &0  &1330\\
NiftyReg-NMI   &83.17 ( 2.76) &0 & 59.65 (7.62) &0  &143\\
%NiftyReg-lncc-20  &0.8196 & 0.5814\\
NiftyReg-LNCC  &83.35 (2.70) & 0  & 67.92 (5.24) &203.3 [35.19] &270\\
vSVF-opt & 82.99 (2.68) &0 &67.35 (9.73)  &0  &79\\
VoxelMorph(w/o aff) &71.25 (9.54) &2.72 [1.57] & 46.06 (14.94) &83.0 [18.13] &0.12 \\
VoxelMorph(with aff)&82.54 (2.78) &5.85 [0.59]  &66.08 (5.13) &39.0 [3.31] &0.31\\
AVSM (2-step)   &82.60 (2.73) &0  & 67.59 (4.47) & 5.5 [0.39] &0.62\\
AVSM (3-step)  &82.67 (2.74) &3.4 [0.12] &\textbf{68.40} (4.35) & 14.3 [1.07]& 0.83\\
% AVSM   &0.8245 (0.027) &0.  & \textbf{0.6841} (None) & 177  & 2.7s(2iter),3.4s(6iter)\\
%\bottomrule
\hline
%0.6841 173 with lbgfs setting
\end{tabular}}
\vspace{0.5ex}
\caption{\label{tab:per_compare} Dice scores (standard deviation) of different registration methods for longitudinal and cross-subject \mnh{registrations} on the OAI dataset. \textit{Affine-opt} and \textit{vSVF-opt} refer to \mnh{optimization-based} multi-scale affine and vSVF \mn{registrations}. \mnh{\textit{AVSM} ($n$-step) refers} to a seven-step affine network and \mn{an} $n$-step vSVF model. \textit{Folds} \mnh{($|\{x:J_{\phi}(x)<0\}|$)} refers to the average number of folds and corresponding absolute Jacobi determinant value in square \mnh{brackets}; \textit{Time} refers to \mnh{the} average \mnh{time} per image registration. 
}
\end{table}

\renewcommand{\tabcolsep}{3pt}
\begin{table*}[!t]
  \begin{center}
  \scalebox{0.7}{
  \begin{tabular}{|l|c|c|c|c|c|c|c|c|c|c|c|c|}
    \hline
    ~\textbf{Method}~ & ~\textbf{Af-Reg}~ & ~\textbf{Af-Multi}~ & ~\textbf{Af-Sym}~&~\textbf{Af-MK}~&~\textbf{vSVF}~&~\textbf{vSVF-MK}~ &~\textbf{vSVF-Multi} ~ & ~\textbf{vSVF-Sym} ~ & ~\textbf{Longitudinal} ~ & ~Better?~&~\textbf{Cross-subject} ~ & ~Better?~ \\
    \hline
     I & &    &   &   & &  &  & &  \textbf{-}  &  & \textbf{-} &  \\
     II& \cmark    &   &   & &  &  &  & & 55.41 & \cmark & 28.68  & \cmark \\
     III & \cmark    & \cmark   &&   & &   &  & & 64.78 &\cmark & 36.31  & \cmark \\
     IV& \cmark &\cmark & \cmark &  & &  && & 68.87  &\cmark & 37.54 & \cmark\\
     
     V&  \cmark& \cmark& \cmark& \cmark &  && &  & 77.75 & \cmark  &   44.58  & \cmark \\
     VI&  \cmark& \cmark& \cmark& & \cmark &  &&& 80.71 &\cmark&59.21  & \cmark \\

     VII&  \cmark& \cmark& \cmark&\cmark & \cmark & \cmark & & & 81.64 & \cmark  &  64.56   & \cmark \\
    
     VIII&  \cmark& \cmark& \cmark&\cmark &\cmark & \cmark & \cmark && 82.81 & \cmark  & 69.08   & \cmark \\
      IV&  \cmark& \cmark& \cmark&\cmark &\cmark & \cmark & \cmark &\cmark& 82.67 & \xmark  & 68.40   & \xmark \\

    \hline
  \end{tabular}}
  \end{center}
  \caption{Ablation study of AVSM using different combinations of methods. \textbf{Af-} and \textbf{vSVF-} separately \mnh{refer} to \mnh{the} affine and \mnh{to the} vSVF related \mnh{methods;} \textit{Reg} refers to \mnh{adding epoch-dependent} regularization\mnh{;} \textit{Multi} refers to multi-step training and testing\mnh{;} \textit{Sym} refers to \mnh{adding the} symmetric loss\mnh{;} \textit{MK} refers to \mnh{using} mk-LNCC as similarity \mnh{measure} (default NCC). Except \mnh{for the last approach which uses} \textit{vSVF-Sym} (last row) \mnh{and encourages symmetric vSVF solutions}, all other \mnh{approaches result in performance improvements.}
  %\mnl{Needs better description and a sentence summarizing what we should see.}\mnl{What is svf-cycle and af-cycle? Is this the same as multi-step and self-iter? Please choose a consistent terminology and use it throughout. Reply: Fixed}
  }
  \label{tab:contributions}
\end{table*}

\begin{figure}[!h]
\centering
\includegraphics[width=0.48\textwidth,height=5cm]{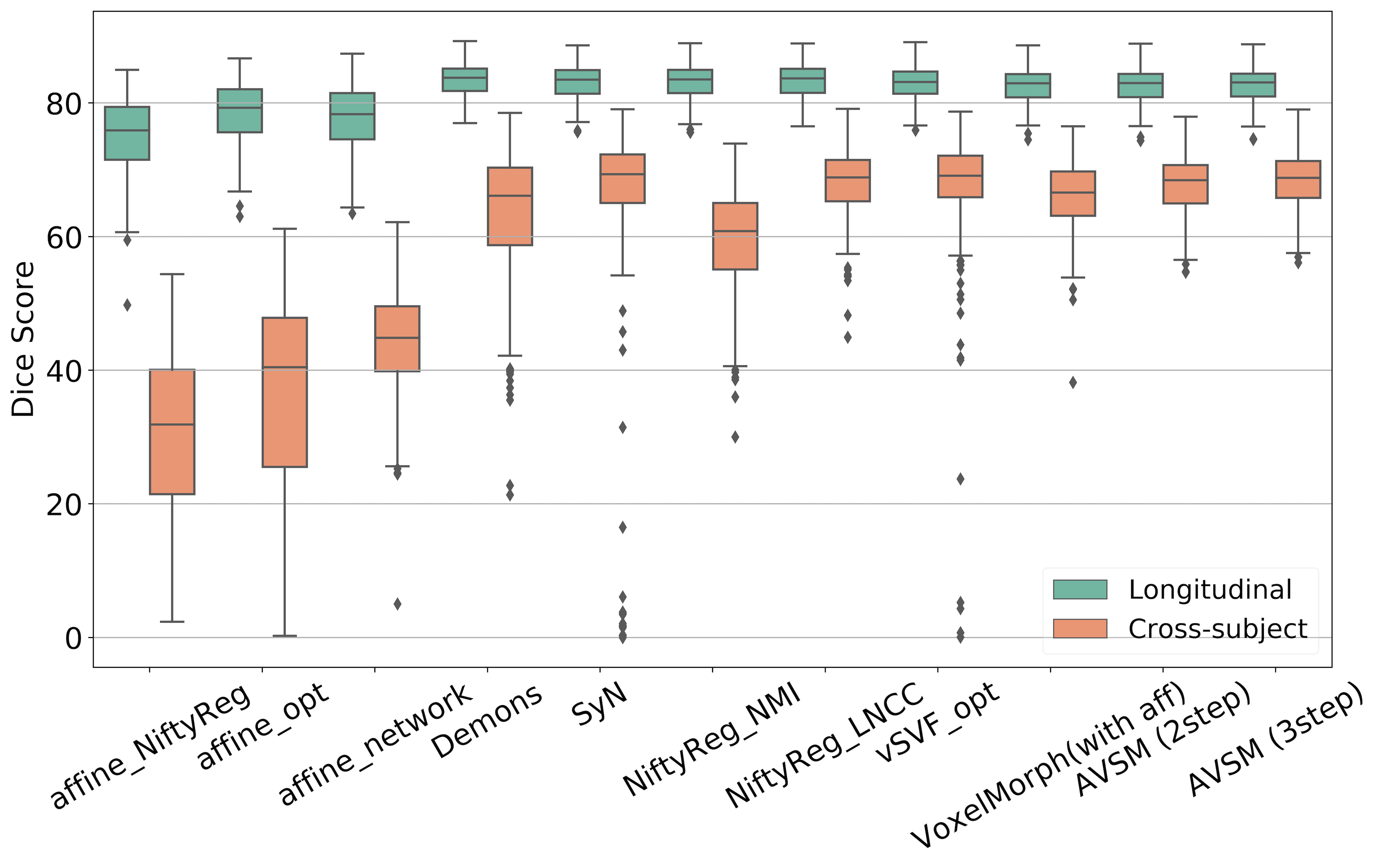}

\caption{\label{fig:bx_intra} Box-plots of the performance of the different registration methods for longitudinal registration\mnh{ (green)} and cross-subject registration (orange). Both AVSM and NiftyReg (LNCC) show high performance and small variance. 
%\xuh{I removed the hspace\{-1cm\}. I think the size is fine, but you may want to enlarge the fonts.}
%\mnl{Add a sentence summarizing what you should see.}\mnl{Add descriptions of methods. What is affine-cycle for example?}\mnl{Are these multi-step results for affine and vSVF?}
}
\end{figure}

Tab.\ref{tab:per_compare} compares the performance of our framework with its corresponding optimization version and public registration tools. Overall, our \mnh{AVSM} framework performs best in cross-subject registration and achieves \mn{slightly better performance than} optimization-based methods\mn{, both for affine and non-parametric registrations. NiftyReg with LNCC shows similar performance.} \mnh{For} longitudinal registration, \mn{AVSM shows good performance, but slightly lower than the optimization-based methods, including vSVF-opt which AVSM is based on.} A possible explanation is that for longitudinal \mn{registrations} deformations are subtle \mn{and source/target image pairs are very similar in appearance. Hence, numerical optimization can very accurately align such image-pairs at convergence.} VoxelMorph runs fastest among all the methods. Without initial affine registration, it unsurprisingly performs poorly. Once the input pair is well pre-aligend,  VoxelMorph shows competitive results for longitudinal registrations, but is outperformed by our approach for the more challenging cross-subject registration.
\mn{To evaluate} the smoothness of the transformation map, we compute the determinant of the Jacobian \mnh{of the estimated map}, \mn{$J_{\phi}(x) := |D{\phi^{-1}}(x)|$},
%,where $x\in R^3$ is the voxel coordinate. 
and count folds defined by \mnh{$|\{x:J_{\phi}(x)<0\}|$ in each image ($192\times192\times80$ voxels in total). We also report the absolute value of the determinant of the Jacobian  in these cases indicating the severity of the fold.} %as well as corresponding $|J_{\phi}(x)<0|$.
Even though the \mn{regularization is} used, \mn{numerical optimization (vSVF-opt) always results in diffeomorphisms}, but very few \mn{folds remain} in AVSM for cross-subject registration. This may be caused by numerical discretization artifacts, by very large predicted momenta, or by inaccuracies of the predictions with respect to the numerical optimization \mn{results.}
%We find the smallest smoothing kernel,\{ 0.05\}, in our multi-Gaussian kernel brings those folds, which is a compromise of sharp displacement.
%\mnh{For affine registration, the multi-step affine network} significantly outperforms other affine methods \mnh{in cross-subject registration}.
%\mnl{How do I see this? And where do I see this in this Figure? Is the correct Figure refefenced here? Reply: Fixed}
\mn{Fig.~\ref{fig:bx_intra}} shows the corresponding boxplot results. AVSM achieves small variance and high performance in both \mnh{registration} tasks and exhibits less registration failures (outliers). \mnh{As AVSM only requires one forward pass to complete both the affine and the vSVF registration, it is much faster than using iterative numerical optimization.}

\begin{figure}[!ht]
	\centering
	%\hspace{-1.cm}
	\includegraphics[width=0.4\textwidth]{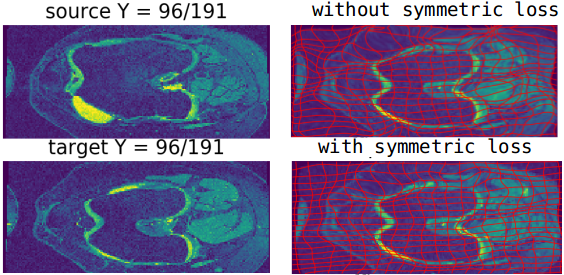}
	
	\caption{\label{fig:sym} Illustration of symmetric loss \mnh{for AVSM}. The left column \mnh{shows the source and target images}. The right column \mnh{shows} the warped image \mn{from a network} trained with and without symmetric loss. The deformation with symmetric loss is smoother.}
\end{figure}

Tab. \ref{tab:contributions} \mnh{shows results} for an ablation study on AVSM. For the affine part, it is difficult to train the single-step affine network without the regularization term. \mnh{Hence, registrations fail.} Introducing multi-step and inverse consistency boosts the affine performance. Compared with \mnh{using} NCC as similarity measure, our \mnh{implementation of mk-LNCC improves results greatly}. In the following vSVF part, we observe a large difference between methods IV and VI, illustrating that vSVF registration results in large improvements. \mnh{Adding mk-LNCC} and multi-step training in \mnh{methods} VII and VIII \mnh{further improves performance}. The exception is the vSVF symmetry \mn{loss} which slightly worsens the performance for both longitudinal and cross-subject registration, but results in good symmetry measures (see Fig.~\ref{fig:sym}).

\begin{figure}[!ht]
\centering
\includegraphics[width=0.45\textwidth]{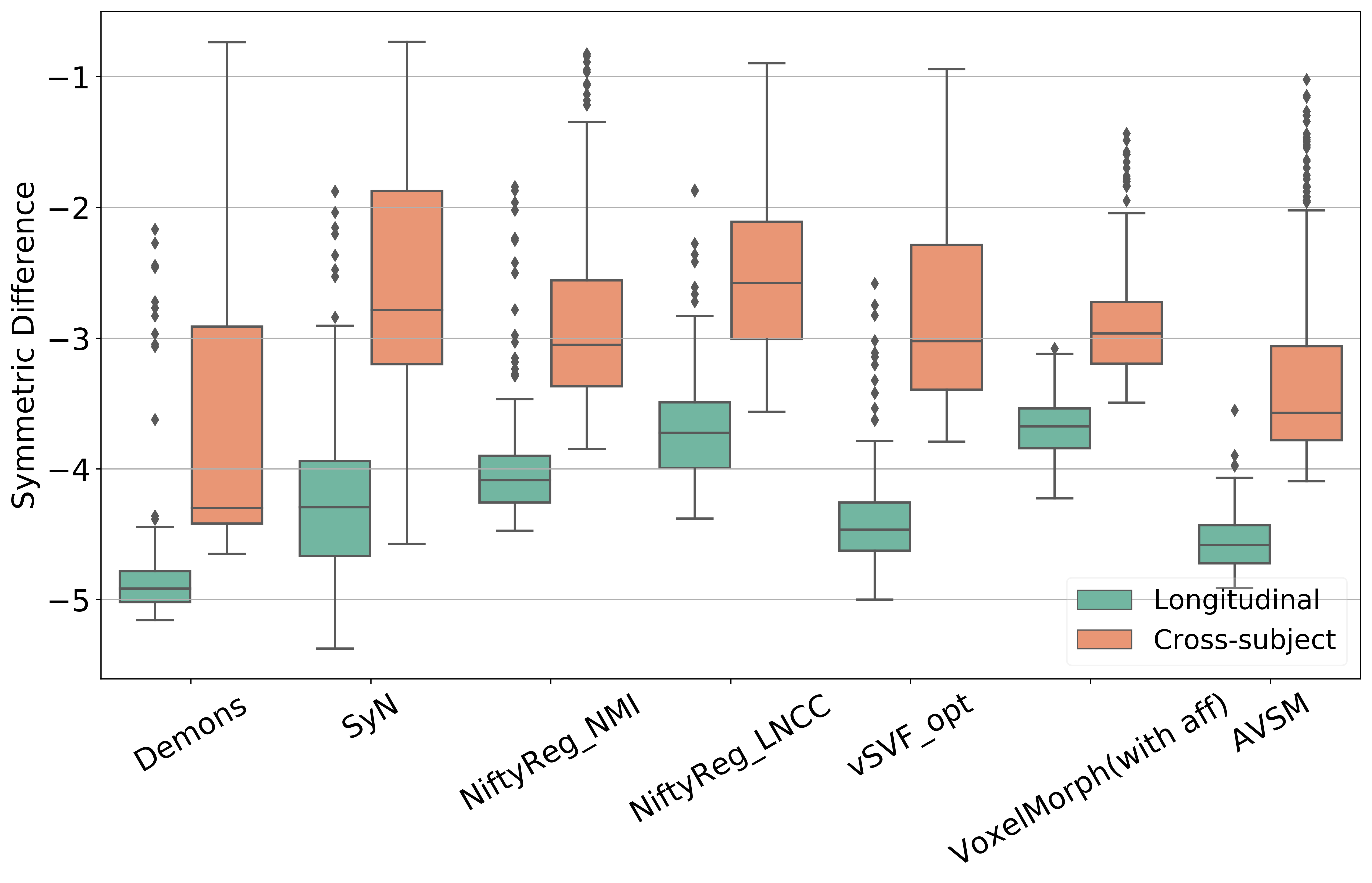}

\caption{\label{fig:sym_comp} Box-plots of the symmetry evaluation (the lower the better) of different registration methods for longitudinal registration\mnh{ (green)} and cross-subject registration (orange). \mnh{AVSM (tested with two-step vSVF) shows good results.}
%\xuh{I removed the hspace\{-1cm\}. I think the size is fine, but you may want to enlarge the fonts.}
}
\end{figure}

We still retain the symmetric loss as it helps the network to converge to solutions with smoother maps as shown in \mn{Fig.~\ref{fig:sym}}. Instead of using larger Gaussian kernels, which can remove \mnh{local displacements}, penalizing asymmetry helps regularize the deformation without smoothing the map too much and without sacrificing too much performance. To numerically evaluate the symmetry, we compute \mnh{$\ln(\frac{1}{|V|}\|\Phi^{-1}\circ(\Phi^{ts})^{-1}- id\|_2^2)$ for all} registration methods, where $V$ refers to the volume size and $\Phi$ the map \mnh{obtained via composition of the affine and the deformable transforms.} \mnh{Since different methods treat boundaries differently, we only evaluate this measure in the interior of the image volume (10 voxels away from the boundary).}
%\xuh{(What do you mean by 'treated differently among methods? shrink, shrank, shrunk...)}
\mnh{Fig.~\ref{fig:sym_comp} shows the results. AVSM obtains low values for both registration tasks, \mn{confirming} its good symmetry properties. Both the Demons and SyN also encourage symmetry, but only AVSM shows a nice compromise between accuracy and symmetry.}
%\mnl{It also does not improve for the cross-subject registrations does it?}. 
%This may be due to the relatively small deformations after affine alignment, which is can not be penalized a lot compared with boundary distortions.\mnl{What do you mean by the second part of this sentence? Could not follow, so did not edit.} \mnl{If you claim improvements here this should be tested with a statistical test. Ideally some form of rank-order test (not t-test). Similarity can be tested for with two one-sided tests as in Xiao's paper. Maybe talk to Zhipeng or Xu, they have both done this before.}

\begin{figure}[!h]
\centering
\includegraphics[width=0.5\textwidth,height=3.2cm]{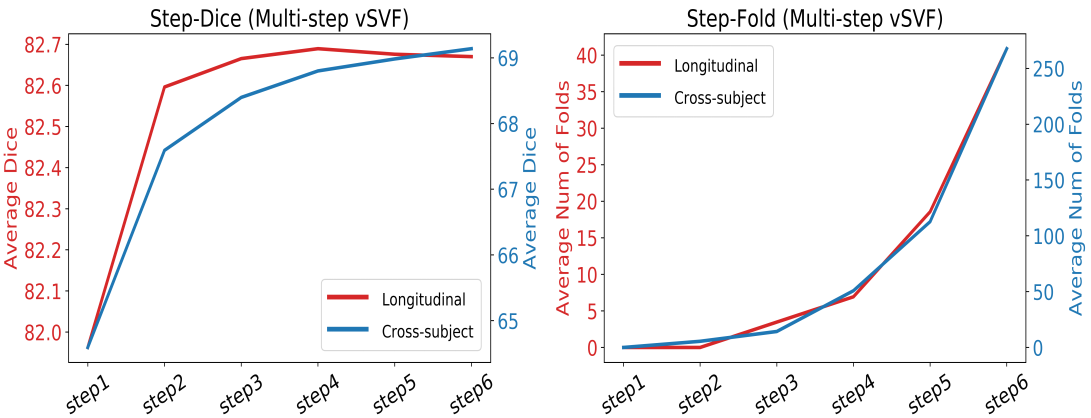}
\caption{\label{fig:dice_jacobi} \mnh{Multi-step vSVF registration results for two-step vSVF training.} \mnh{Performance increases with steps (left), but the number of folds also increases (right).}% The left plot shows the average dice score over the steps. The right plot shows the average folds number over the steps. 
%\xuh{I removed the hspace\{-1cm\}. I think the size is fine, but you may want to enlarge the fonts.}
}
\end{figure}

\mnh{Fig.~}\ref{fig:dice_jacobi} \mnh{shows the average Dice sores over the number of test iteration steps of vSVF.} The model is trained \mnh{using a} two-step vSVF. It can be \mnh{observed that iterating the} model for more than two steps can increase performance \mnh{as these iterations result in registration refinements.} However, the average number of folds also increases, \mnh{mostly at boundary regions and in regions of anatomical inconsistencies. Examples are shown in the supplementary material.}

% %\mnh{This entire discussion section is the meat of your work. It needs to be substantially improved and any claims you made earlier in the text need to be justified here with experimental results. It would also be good to report some measures of registration regularity, e.g., statistics of values of the determinant of the Jacobian of the transformation.}

% \begin{figure}[!h]
% \centering
% \hspace{-1.cm}
% \includegraphics[width=0.4\textwidth]{Figs/vSVF_iter.png}

% \caption{\label{fig:vSVF_iter} \mns{Illustration of symmetric loss in vSVF. The left column is the souce and target image. The right column is the warped image trained with and without vSVF symmetric loss. The deformation (red) with symmetric loss is smoother.}\mnl{Wrong caption.}}
% \end{figure}

\section{Conclusions and Future Work}
\label{sec:conclusions_and_future_work}

\mnh{We introduced an end-to-end 3D image registration approach (AVSM) consisting of a multi-step affine network and a deformable registration network using a momentum-based SVF algorithm. AVSM outputs a transformation map which includes an affine pre-registration {\it and} a vSVF non-parametric deformation in a single forward pass. Our results on cross-subject and longitudinal registration of knee MR images show that our method achieves comparable \mn{and sometimes better} performance to popular registration tools with a \mn{dramatically reduced computation time} and with excellent deformation regularity and symmetry.} \mnh{Future work will focus on also learning regularizers and evaluations on other registration tasks, \eg in the brain and the lung.}

%  Currently, our work \mnh{focuses} on \mnh{the} OAI \mnh{dataset}, but for \mnh{future} work, we are interested in fast transfer learning 
%\xuh{(Should you cite one paper here?)} 
%\mnh{methods} to generalize our framework \mnh{to} other registration tasks.  %Notably, at the same time we also encourage inverse-consistency\mnl{Would be good to quantify this somehow Reply: This part need to be extended}.\mnl{This conclusion and future work section needs to be expanded once we have all the results and potential future work needs to be included.} 

% \mnl{We will need to improve this section once we have all the results.}

% \mnr{Check the bibiliography for proper capitalization, e.g.,, MR instead of mr.}
\noindent
\textbf{Acknowledgements}: Research reported in this publication was supported by the National Institutes of Health (NIH) and the National Science Foundation (NSF) under award numbers NSF EECS1711776 and NIH 1R01AR072013. The content is solely the responsibility of the authors and does not necessarily represent the official views of the NIH or the NSF.
\appendix

\section{Supplementary material}

\mn{
	This supplementary material provides additional details illustrating the proposed approach. Specifically, Sec.~\ref{sec:affine_regularization} describes how the affine training is regularized in an epoch-dependent way. Sec.~\ref{sec:dice_over_steps_affine} shows registration performance for different numbers of steps for the affine registration-part of the network. Sec.~\ref{sec:structure_of_momentum_generation_network} details the structure of the momentum generation network. Lastly, Sec.~\ref{sec:visualization} shows additional registration examples.
}

\subsection{Affine regularization factor}
\label{sec:affine_regularization}
\begin{figure}[!h]
	\centering
	\includegraphics[width=0.4\textwidth]{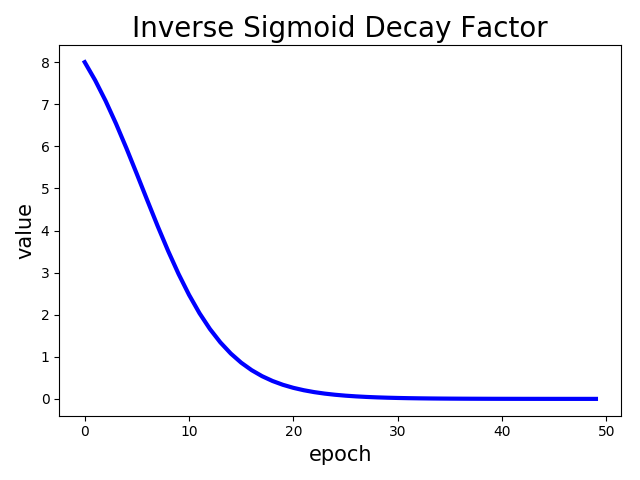}
	\caption{\label{fig:factor} \mn{Graph of the} affine regularization factor. \mn{Its} value decays to zero over the epochs.}
\end{figure}

\mn{To help with convergence of the affine registration network, we use an epoch-dependent regularization factor, which discourages large transformations at the start of the training. Specifically, we define this epoch-dependent regularization factor as}
\begin{equation}
\lambda_{ar} := \frac{C_{ar} K_{ar}}{K_{ar}+e^{n/K_{ar}}},
\label{eq:regularization_penalty_weights}
\end{equation}
where $C_{ar}$ is a constant, $K_{ar}$ controls the decay rate, and $n$ is the epoch count.
\mn{Fig.~\ref{fig:factor} shows the value of $\lambda_{ar}$ plotted over the epochs. As the value decays to zero with the epochs, its influence on the training becomes negligible.} \mn{For both} longitudinal and cross-subject experiments, $K_{ar}$ is set to 4 and  $C_{ar}$ is set to 10. 
%In our experiments, we use the constant $\lambda_{ar}=C_{ar}$ in the first 5 epoch and then the decay function, which means $n:= \max(0,n-5)$.

\subsection{Dice over steps in Multi-step Affine Network}
\label{sec:dice_over_steps_affine}

\begin{figure}[]
	\centering
	\includegraphics[width=0.4\textwidth]{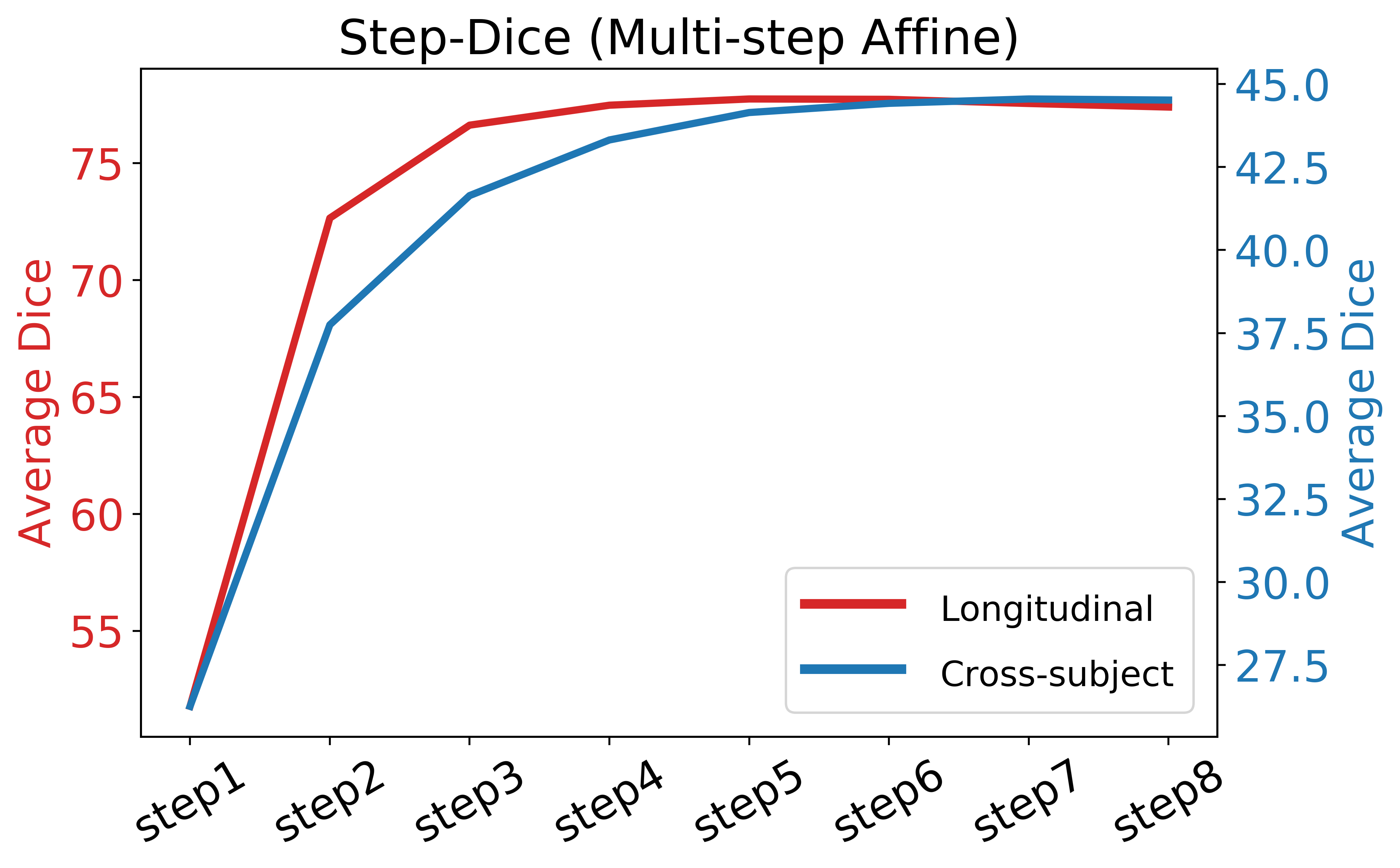}
	\caption{\label{fig:step_af.png} Multi-step Affine registration results over iteration steps. The affine network is trained \mn{using} three steps for longitudinal registration ({\color{red}red}) and five steps for cross-subject registration ({\color{blue}blue}). Performance increases with steps and finally saturates.}
\end{figure}

\mn{The main manuscript shows the average Dice scores over the number of test iteration steps for the vSVF registration component.} For completeness\mn{, Fig.~\ref{fig:step_af.png} shows} the average Dice scores over the number of steps \mn{for} the affine network. The model is trained using a three-step affine network \mn{for longitudinal registrations and using five steps for cross-subject registration.} Similar to the vSVF registration, it can be observed that \mn{model performance improves with large number of steps and saturates at a high performance level.}

\subsection{Structure of Momentum Generation Network}
\label{sec:structure_of_momentum_generation_network}

\begin{figure*}[]
	\centering
	\includegraphics[width=1\textwidth]{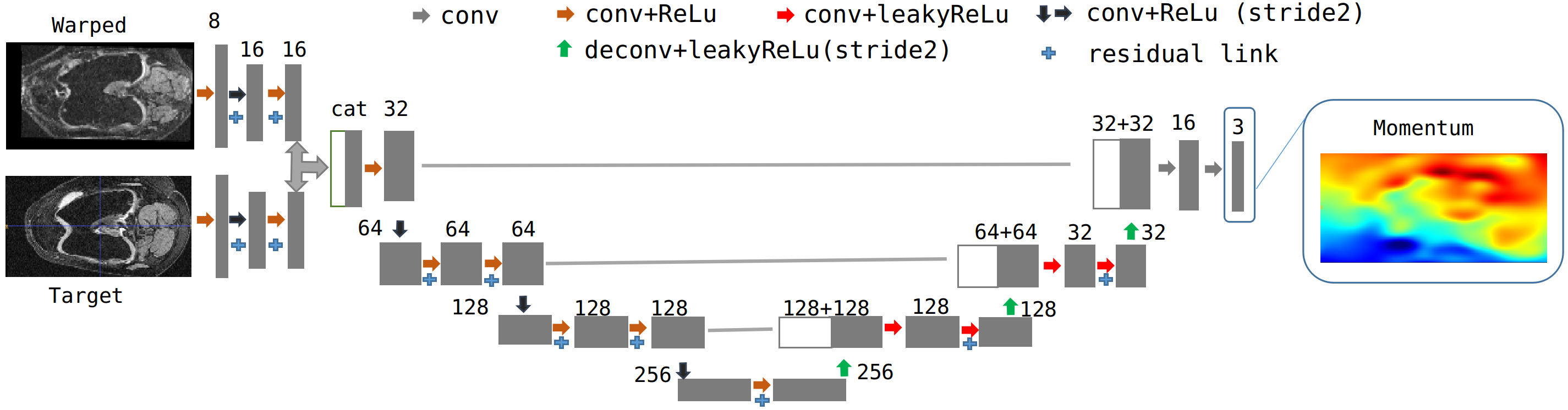}
	\caption{\label{fig:unet} Illustration of the structure of Momentum Generation Network. It follows the structure of the U-net but the last level decoder is removed. 
	}
\end{figure*}

\mn{As the network structure itself is not the main contribution of our work, we do not describe it in detail in the main manuscript. For completeness, we describe the architecture here.} Fig.~\ref{fig:unet} shows the structure of the Momentum Generation Network. It takes a pair of images as the input and outputs a low-resolution initial momentum. \mn{We use a} four-level U-net ~\cite{ronneberger2015u,milletari2016v} with residual \mn{links}, but \mn{remove the last decoder level to output the low-resolution momentum.} \mn{As the momentum can be positive or negative, no activation function (\eg ReLu ~\cite{nair2010rectified} or leakyRelu ~\cite{maas2013rectifier}) is used after the last two convolutional layers, which output the momentum.}

\subsection{Visualization}
\label{sec:visualization}

\mn{To provide more insight into the registration behavior of our network, we visualize results illustrating deformation folds, results for different steps in the multi-step approach, and additional examples. Specifically, we show the following:}

\begin{figure*}[]
	\centering
	\includegraphics[width=0.95\textwidth]{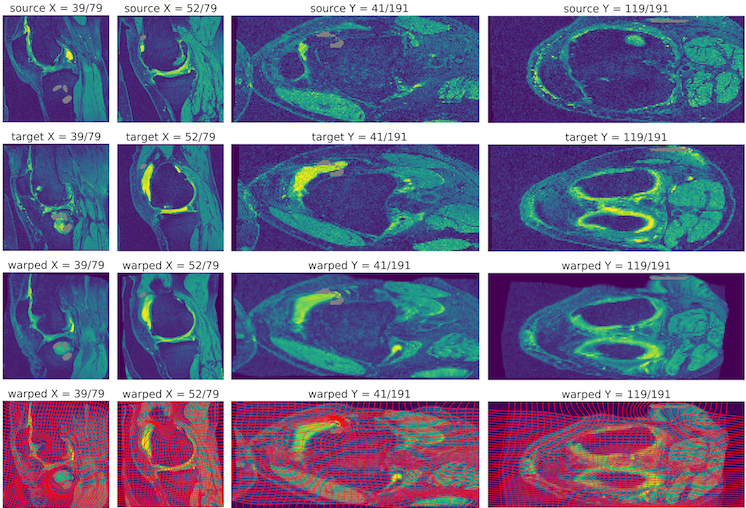}
	\caption{\label{fig:folds} Examples of folds produced by \mn{a} six-step vSVF (trained \mn{using a} two-step vSVF). Each column refers to an example registration case. From top to bottom \mn{source, target, warped image by AVSM and warped image with deformation grid (visualizing $\Phi^{-1}$) are shown}. Folds are shown in \textit{gray}. From left to right, the first three columns refer to cases with anatomical inconsistency and the last column refers to a case where the folds occur at the boundary. 
	}
\end{figure*}

\begin{figure*}[t]
	\centering
	\includegraphics[width=1\textwidth]{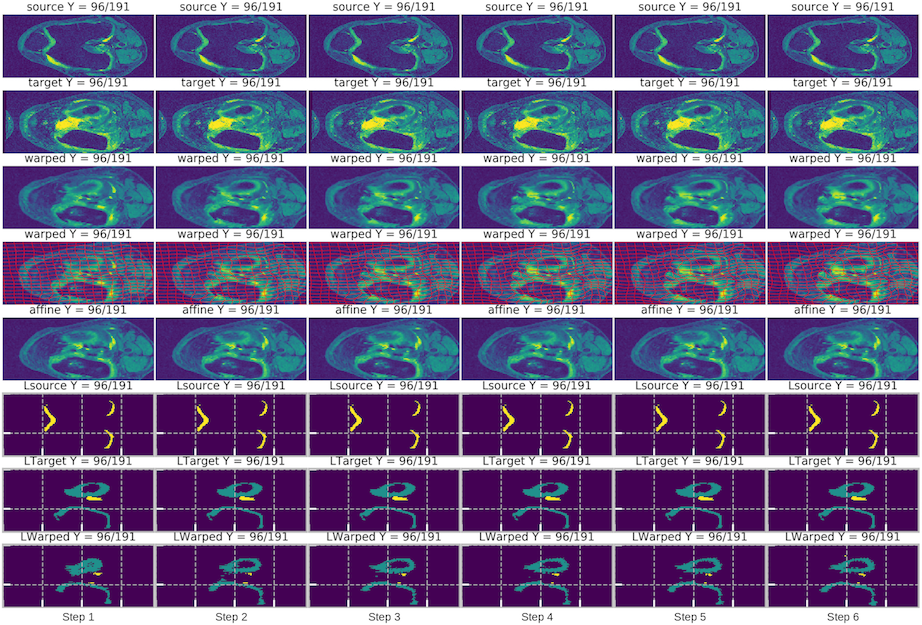}
	\caption{\label{fig:steps} Illustration of the results of \textit{one registration case (with six steps)} by AVSM (trained \mn{using a} two-step vSVF). From left to right, each column \mn{shows results for different steps.} The first five rows refer to source, target, warped image by AVSM, warped image with deformation grid (visualizing $\Phi^{-1}$) \mn{and} warped image by \mn{the multi-step affine network} respectively\mn{. The last three rows} show the source label, target label and warped label \mn{for the  AVSM result}. The transformation map gets refined over \mn{the} six steps and the registration result \mn{improves as indicated by a better correspondence between the target label and the warped label images (last two rows)}. }
\end{figure*}

\begin{figure*}[!t]
	\centering
	\includegraphics[width=1\textwidth]{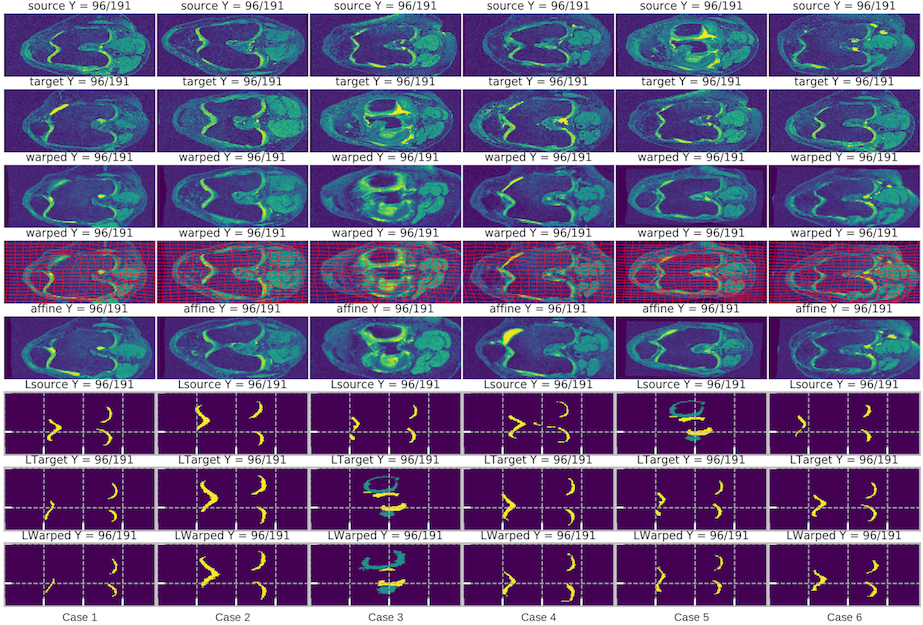}
	\caption{\label{fig:cases} Illustration of results of \textit{six registration cases} by AVSM. Each column refers to an example registration case. The first five rows refer to source, target, warped image by AVSM, warped image with deformation grid (visualizing $\Phi^{-1}$) \mn{and} warped image by \mn{the multi-step affine network} respectively\mn{. The last three rows show the source label, target label and warped label by AVSM}. There is high similarity between the warped and the target images and the deformations are smooth,\mn{ illustrating the good registration performance of our proposed AVSM approach}.}
\end{figure*}

\begin{itemize}
	\item \textit{Folds}: To better visualize the folds produced by the multi-step vSVF, we report the registration results, shown in Fig.~\ref{fig:folds}, from the six-step vSVF. These folds mostly occur at regions of anatomical inconsistency or \mn{at the image boundary where map interpolation artifacts may influence the solution.} In these regions, very large \mn{momentum values may be predicted which can result in folds due to discretization artifacts when integrating the advection equation.}
	
	\item \textit{Multi-step in vSVF registration}: 
	Fig.~\ref{fig:steps} shows the registration results over the steps of \mn{the} vSVF. Although \mn{folds may result from the multi-step strategy in some very large deformation cases,} the transformation maps are \mn{largely well regularized.} We observe that the registration results \mn{improve over the steps.}
	
	\item \textit{More AVSM examples}: \mn{Fig.~\ref{fig:cases} shows additional AVSM registration results.} It can be observed that AVSM achieves \mn{good} registration results with smooth transformation maps \mn{for cases with} large and small \mn{deformations.}
\end{itemize}

{\small
\bibliographystyle{ieee_fullname}
\bibliography{main}
}
\end{document}